\documentclass[10pt,twocolumn]{article}
\usepackage{graphicx}
\usepackage{amsmath,amssymb,times,cite}
\usepackage{booktabs}
\usepackage[dvipsnames]{xcolor}

\usepackage[hidelinks, bookmarks=false]{hyperref}
\usepackage{hyperref}
\hypersetup{
	colorlinks = true,
	citecolor = red,
	linkbordercolor = white,
}

\newcommand{\bg}[1]{\boldsymbol{#1}} 
\newcommand{\bm}[1]{\mathbf{#1}} 

\newcommand\T{{\mathpalette\raiseT\intercal}}
\newcommand\raiseT[2]{%
\setbox0\hbox{$#1{#2}$}\raise\dp0\box0}

\usepackage{pifont}
\newcommand{\cmark}{\ding{51}}%
\newcommand{\xmark}{\textcolor{lightgray}{\ding{55}}}%

\title{\Large\textbf{A Hybrid Kolmogorov-Arnold Network for Medical Image Segmentation}}
\author{Deep Bhattacharyya, Ali Ayub, and A. Ben Hamza\\
Concordia Institute for Information Systems Engineering\\
Concordia University, Montreal, QC, Canada
}
\date{}

\begin{document}
\maketitle

\begin{abstract}
Medical image segmentation plays a vital role in diagnosis and treatment planning, but remains challenging due to the inherent complexity and variability of medical images, especially in capturing non-linear relationships within the data. We propose U-KABS, a novel hybrid framework that integrates the expressive power of Kolmogorov-Arnold Networks (KANs) with a U-shaped encoder-decoder architecture to enhance segmentation performance. The U-KABS model combines the convolutional and squeeze-and-excitation stage, which enhances channel-wise feature representations, and the KAN Bernstein Spline (KABS) stage, which employs learnable activation functions based on Bernstein polynomials and B-splines. This hybrid design leverages the global smoothness of Bernstein polynomials and the local adaptability of B-splines, enabling the model to effectively capture both broad contextual trends and fine-grained patterns critical for delineating complex structures in medical images. Skip connections between encoder and decoder layers support effective multi-scale feature fusion and preserve spatial details. Evaluated across diverse medical imaging benchmark datasets, U-KABS demonstrates superior performance compared to strong baselines, particularly in segmenting complex anatomical structures.
\end{abstract}

\bigskip
\noindent\textbf{Keywords}:\, Medical image segmentation; Kolmogorov-Arnold network; encoder-decoder architecture; Bernstein polynomials.

\section{Introduction}\label{sec:intro}
The aim of medical image segmentation is to partition medical images, such as ultrasound, MRI, or dermoscopic images, into meaningful regions corresponding to anatomical structures or pathological areas, assigning each pixel a class label to delineate organs, tissues, or lesions. This task is fundamental to computer-aided diagnosis, enabling precise identification of regions of interest for clinical analysis, and disease monitoring. Recent advances in deep learning have driven significant progress in medical image segmentation~\cite{Xu2023DUCT}, with several approaches emerging to tackle its challenges. Fully Convolutional Network (FCN)-based methods such as U-Net~\cite{Ronneberger2015} and its variants, UNet++~\cite{Zhou2018}, Attention U-Net~\cite{Oktay2018}, UNet3+~\cite{Huang2020}, 3D UNet~\cite{Cicek2016}, V-Net~\cite{Milletari2016}, and Ki-UNet~\cite{Valanarasu2020a, Valanarasu2020b}, leverage convolutional layers to capture local spatial patterns, achieving robust performance. However, they often struggle with global context modeling, leading to over-segmentation or diffused boundaries in noisy images. In medical segmentation, we use the term ``global context'' to refer to (i) long-range boundary coherence across distant pixels, (ii) topology preservation (avoiding spurious holes/mergers), and (iii) global shape regularity consistent with anatomical structures.

To address these challenges, Transformer-based U-shaped architectures, such as MedT~\cite{Valanarasu2021} and Swin-Unet~\cite{Cao2021}, have been developed to capture global spatial patterns through self-attention mechanisms. Yet, their quadratic computational complexity with respect to input length and reliance on large datasets limit their applicability in resource-constrained clinical settings or data-scarce scenarios. Multi-Layer Perceptron (MLP)-based methods, such as UNeXt~\cite{Valanarasu2022} and MALUNet~\cite{Ruan2022}, prioritize efficiency, but their linear transformations often lack the expressiveness needed for complex lesion boundaries, resulting in under-segmentation. More recently, Kolmogorov-Arnold Networks (KANs)~\cite{Liu2024KAN} have emerged as a promising alternative to MLPs, leveraging learnable activation functions to enhance interpretability and expressiveness. Building on this, U-KAN~\cite{li2024ukan} incorporates B-spline activation functions within a U-shaped architecture to effectively model local spatial patterns, achieving competitive results across various medical image segmentation datasets. However, it struggles to capture global context.

The aforementioned limitations highlight the need for a segmentation framework that combines computational efficiency, robust global-local feature modeling, and adaptability to diverse medical imaging modalities. To this end, we propose a novel hybrid framework, U-KABS, which integrates a U-shaped encoder-decoder architecture with a KAN Bernstein spline (KABS) block as a core component, combining the global smoothness of Bernstein polynomials and the local adaptability of B-splines. This hybrid design enables our model to effectively capture both global contextual information and fine-grained local details, which are critical for the accurate segmentation of complex anatomical structures in medical images. Our main contributions can be summarized as follows:
\begin{itemize}
\item We propose a novel hybrid Kolmogorov-Arnold network, integrating a U-shaped encoder-decoder architecture, and leveraging skip connections and squeeze-and-excitation mechanisms to capture hierarchical features and prioritize salient regions.
\item We introduce the KAN Bernstein spline block, combining Bernstein polynomials and B-splines to balance global smoothness and local adaptability.
\item We conduct extensive evaluations on four benchmark datasets with varying characteristics, demonstrating superior performance over state-of-the-art baselines and robustness to blurry boundaries, and multi-class tasks.
\end{itemize}

\section{Related Work}
\noindent\textbf{FCN-based Methods.}\quad The U-Net architecture~\cite{Ronneberger2015} is a foundational FCN-based model for medical image segmentation, featuring a symmetric encoder-decoder structure with skip connections to preserve spatial details across scales. Building on U-Net's success, several variants have been developed to further improve segmentation performance. For instance, UNet++~\cite{Zhou2018} enhances U-Net by incorporating nested and dense skip connections, which improve feature aggregation. To address the need for focusing on salient regions, attention-based FCN methods have emerged. Attention U-Net~\cite{Oktay2018} integrates attention mechanisms to emphasize relevant features. Similarly, MALUNet \cite{Ruan2022} employs multi-scale attention to capture features at varying resolutions. While FCN-based methods are effective for local feature learning, they often struggle to model long-range dependencies, limiting their ability to capture global contextual information in complex medical images.

\medskip\noindent\textbf{Transformer- and MLP-based Methods.}\quad Drawing inspiration from Vision Transformers~\cite{Dosovitskiy2020}, Transformer-based models have been recently adapted for medical image segmentation to overcome the limitations of FCN-based approaches in capturing global dependencies. TransUNet~\cite{Chen2024TransUNet} integrates Transformer blocks with U-Net architecture, leveraging self-attention and cross-attention mechanisms to enhance performance. MedT~\cite{Valanarasu2021} employs a dual-branch encoder to learn fine-grained spatial details. Swin-Unet~\cite{Cao2021} utilizes shifted window-based self-attention to model long-range dependencies. UNETR~\cite{unetr} introduces a Transformer-based architecture that reformulates volumetric medical image segmentation as a 1D sequence-to-sequence task. WRANet~\cite{Zhao2023wranet} presents an enhanced U-Net architecture that replaces conventional pooling with discrete wavelet transform-based downsampling to suppress high-frequency noise while preserving salient structural information, and integrates residual attention modules to mitigate feature loss and gradient degradation. DECSTNet~\cite{Zhao2025DECSTNet} proposes a dual-encoder network that combines CNNs with a cross-shaped window transformer to capture multi-scale local features and long-range contextual dependencies for medical image segmentation, particularly for intracranial aneurysms, by integrating adaptive dynamic feature fusion and lightweight cross-shaped attention. While Transformer-based methods excel in capturing global context, they often require substantial computational resources and large annotated datasets. To address this computational overhead, lightweight MLP-based methods have gained attention for their efficiency. For instance, UNeXt~\cite{Valanarasu2022} introduces tokenized MLP blocks to capture local and global dependencies using a shifted MLP block to handle spatial relationships across image dimensions. Rolling-Unet~\cite{Liu2024} introduces the Rolling-MLP block to capture long-distance dependencies in multiple directions while preserving local features through a U-Net architecture. These MLP-based approaches offer a balance between performance and computational efficiency, but lack the expressive power needed for complex medical imaging tasks.

\medskip\noindent\textbf{KAN-based Methods.}\quad KANs~\cite{Liu2024KAN} have emerged as a compelling alternative to MLPs, leveraging their ability to approximate multivariate functions through compositions of learnable univariate functions. More recently, U-KAN \cite{li2024ukan}, which builds on the U-Net backbone by incorporating tokenized KAN blocks with B-spline activation functions, has demonstrated improved performance in medical image segmentation. Similarly, ResU-KAN~\cite{wang2025resu} integrates a residual convolutional attention and an atrous spatial pyramid pooling module into U-KAN in an effort to capture long-distance dependencies and expand expand the model's receptive field. However, their reliance on B-splines constrains the model's ability to effectively balance global context with local feature representations. In contrast, our proposed U-KABS framework integrates both Bernstein polynomials and B-splines as learnable activation functions, combining global smoothness with local adaptability, and offering a novel design that achieves superior segmentation performance while maintaining computational efficiency. By integrating this hybrid approach within a U-Net-inspired architecture and leveraging squeeze-and-excitation mechanisms, U-KABS provides a flexible and expressive model well-suited to the complexities of medical image segmentation.

\section{Method}
In this section, we begin by outlining the segmentation task, followed by a brief background on standard KANs using splines as basis functions. We then present an overview of our model architecture and introduce its core components.

\begin{figure*}[!htb]
\centering
\includegraphics[width=0.8\linewidth]{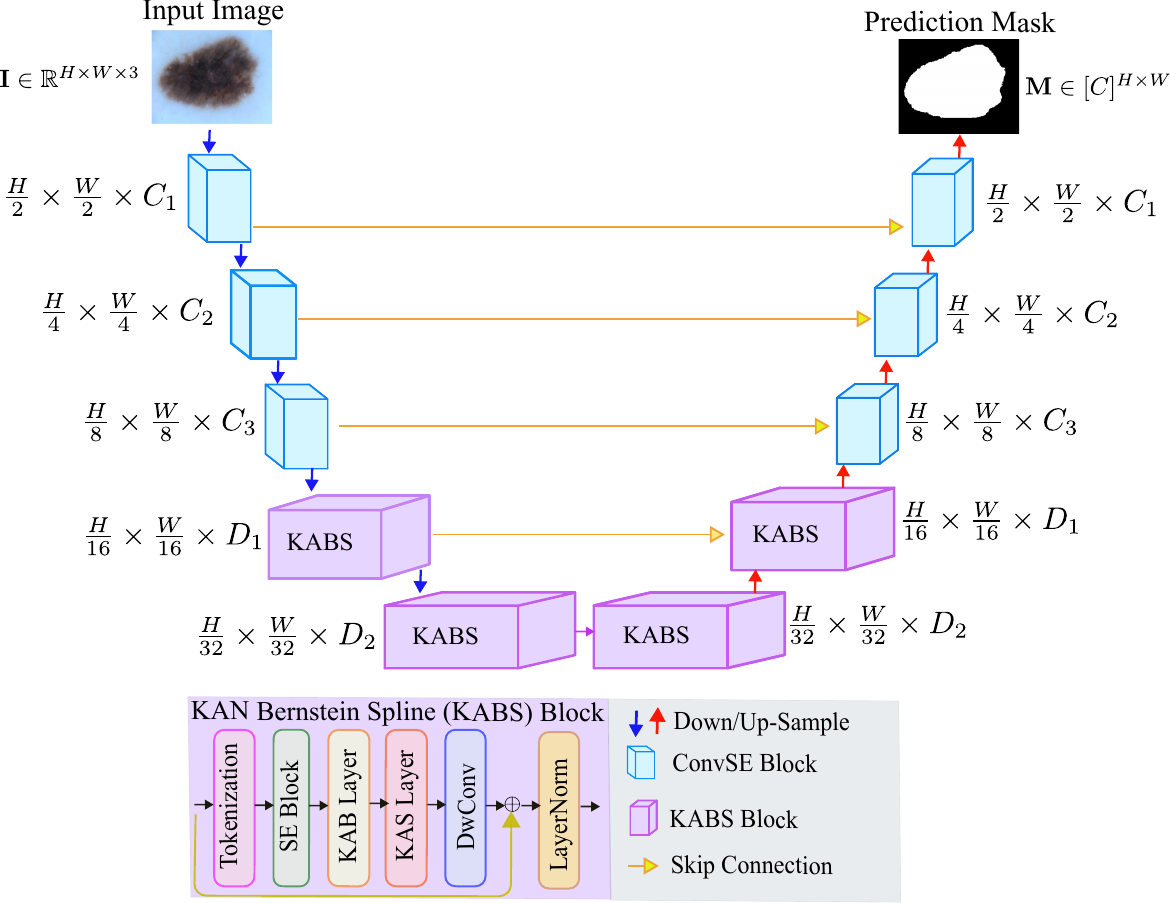}
\caption{Overall architecture of the proposed U-KABS framework. The model follows a symmetrically U-shaped encoder-decoder structure, integrating Convolutional and Squeeze-and-Excitation (ConvSE) and KAN Bernstein Spline (KABS) stages. The encoder, comprising three ConvSE blocks followed by two KABS blocks, progressively downsamples the input image to learn hierarchical features, increasing the number of feature channels while halving spatial resolution at each block. The decoder mirrors the design of the encoder, upsampling features via bilinear interpolation to restore the feature map spatial resolution and generate a pixel-wise segmentation mask.}
\label{Fig:Architecture}
\end{figure*}

\subsection{Preliminaries}
\noindent\textbf{Problem Description.}\quad Let $\bm{I}\in\mathbb{R}^{H\times W\times 3}$ be an input image with height $H$ and width $W$. Our goal is to predict a pixel-wise segmentation mask $\bm{M}\in [C]^{H\times W}$, where $[C]=\{0,\dots,C-1\}$ and $C$ is the number of semantic classes. Each pixel in $\bm{M}$ is assigned an integer label from 0 to $C-1$, indicating the class to which that pixel belongs.

\medskip\noindent\textbf{Kolmogorov-Arnold Networks.}\quad KANs are inspired by the Kolmogorov-Arnold representation theorem~\cite{Braun2009KANTH,Johannes2021KANTH}, which states that any continuous multivariate function on a bounded domain can be represented as a finite composition of continuous univariate functions of the input variables and the binary operation of addition. Specifically, any continuous multivariate function $f:[0,1]^{n}\to\mathbb{R}$ can be expressed as
\begin{equation}
f(\bm{x})=\sum_{q=1}^{2n+1}\Phi_{q}\left(\sum_{p=1}^{n}\phi_{q,p}(x_p)\right),
\end{equation}
where $\bm{x}=(x_1,\dots,x_n)^{\T}\in\mathbb{R}^n$ is the input vector, $\phi_{q,p}: [0,1]\to\mathbb{R}$ and $\Phi_{q}: \mathbb{R}\to\mathbb{R}$ are continuous univariate functions. This theorem guarantees that we can decompose a complex function of multiple variables into a combination of simpler, univariate functions.

A KAN layer is a fundamental building block of KANs~\cite{Liu2024KAN}, and is given by a matrix $\bg{\Phi}=(\phi_{q,p})$ of 1D functions, where each trainable activation function $\phi$ is defined as a weighted combination of a sigmoid linear unit (SiLU) function and a spline function:
\begin{equation}
\phi(x)=w_{b}\text{SiLU}(x) + w_{s}\text{spline}(x),
\label{Eq:spline}
\end{equation}
where $\text{spline}(x)=\sum_{i}c_{i}B_{i}(x)$ is a weighted sum of B-splines basis functions with trainable coefficients $c_{i}$. During training, the weights $w_{b}$ and $w_{s}$ are learned to optimize performance. Given a input feature vector $\bm{x}^{(l)}\in\mathbb{R}^{F_{l}}$, the output of the $l$-th KAN layer is an $F_{l+1}$-dimensional feature vector given by
\begin{equation}
\bm{x}^{(l+1)} = \text{KAN}^{(l)}(\bm{x}^{(l)}).
\end{equation}

\subsection{Architecture Overview}
The overall architecture of the proposed U-KABS framework is illustrated in Figure~\ref{Fig:Architecture}. It follows a symmetric encoder-decoder design, along with skip connections that bridge corresponding blocks, enabling effective feature fusion. U-KABS integrate the expressive capabilities of KANs with convolutional and attention-based squeeze-and-excitation mechanisms, and incorporates two distinct processing stages in both the encoder and decoder: the Convolutional and Squeeze-and-Excitation (ConvSE) stage and the KAN Bernstein Spline (KABS) stage. The ConvSE stage, comprised of three blocks, combines convolutional layers with SE blocks to enhance channel-wise feature representations. This stage is particularly effective for capturing local patterns, such as edges, and recalibrating feature channels to emphasize the most relevant features for segmentation. On the other hand, the KABS stage leverages the principles of KANs, employing learnable activation functions based on Bernstein polynomials and B-splines. These activation functions provide a unique balance between global smoothness and local adaptability, making the KABS stage particularly well-suited for modeling complex patterns in medical images, such as subtle tissue boundaries, tumors, or pathological structures. The decoder mirrors the encoder's structure to generate the segmentation map from the encoded features.

\subsection{Encoder}
The encoder consists of the following two stages:

\smallskip\noindent\textbf{Convolutional and squeeze-and-excitation (ConvSE) stage.}\quad This stage consists of $L$ ConvSE blocks, each of which is comprised of a convolutional layer, a batch normalization (BN) layer, a ReLU activation function, and a squeeze-and-excitation block to enhance channel-wise feature representations. Formally, the process of the ConvSE stage can be expressed as
\begin{equation}
\bm{X}^{(\ell)} = \text{SE}(\text{ReLU}(\text{BN}(\text{Conv}(\bm{X}^{(\ell-1)})))),
\label{Eq:ConvSE}
\end{equation}
where $\bm{X}^{(\ell)}$ is the output feature map of the $\ell$-th ConvSE block for $\ell\in\{1, \dots, L\}$, and $\bm{X}^{(0)} = \bm{I}$ is the input image. A $2\times 2$ max-pooling operation is applied after each block for downsampling. The output feature map $\bm{X}^{(L)}$ of the last ConvSE block serves as input for the first block of the subsequent KAN Bernstein Spline (KABS) stage.

\medskip\noindent\textbf{KAN Bernstein Spline (KABS) stage.}\quad This stage is comprised of $K$ blocks, each of which consists of a tokenization layer, a squeeze-and-excitation block, two KAN layers with distinct basis functions as learnable activation functions, a depth-wise convolution layer, a residual connection, and a layer normalization, as illustrated in Figure~\ref{Fig:Architecture}.

\smallskip\noindent\textit{Tokenization and squeeze-and-excitation.}\quad The feature map $\bm{X}^{(L)}$ of the $L$-th block of the ConvSE stage is converted into a sequence of tokens, yielding an intermediate feature map $\bm{Z}^{(0)}$. This tokenization process follows the Vision Transformer methodology~\cite{Dosovitskiy2020}, where the input feature map is partitioned into a sequence of flattened patches, and projected into a latent embedding space using a trainable linear projection. The features are then processed through a squeeze-and-excitation block before being fed into two subsequent KAN layers with different basis functions for further feature enhancement.

\smallskip\noindent\textit{KAN Bernstein (KAB) Layer.}\quad We approximate each trainable activation function of the KAB layer over the interval $[0,1]$ using a weighted combination of a SiLU function and a Bernstein polynomial function:
\begin{equation}
\tilde{\phi}(z)=\tilde{w}_{b}\text{SiLU}(z) + \tilde{w}_{s}\text{Bernstein}(z),
\label{Eq:Bernstein}
\end{equation}
where
\begin{equation}
\begin{split}
\text{Bernstein}(z) & = \sum_{r=0}^{R} \theta_{r} P_{r,R}(z) \\
                    & = \sum_{r=0}^{R} \theta_{r} {R\choose r}z^{r}(1-z)^{R-r}
\end{split}
\end{equation}
is a weighted sum of Bernstein basis polynomials $P_{r,R}$ of degree $R$, and $\theta_{r}$ are the learnable Bernstein coefficients. Bernstein-based KAN offers global, smooth function approximation, while being memory efficient due to the recurrence avoiding explicit storage of all polynomial terms. Before evaluating the Bernstein basis, token features $z$ are mapped to $[0,1]$ using per-batch min-max normalization with a small $\epsilon$ for numerical stability:
\begin{equation}
\hat{z}=\mathrm{clip}\!\left(\frac{z-\min(z)}{\max(z)-\min(z)+\epsilon},\,0,\,1\right), \quad \epsilon=10^{-6}.
\end{equation}
We then feed $\hat{z}\in [0,1]$ to the Bernstein basis. This procedure preserves relative ordering and guarantees bounded inputs for stable polynomial evaluation. In practice, a Bernstein order of 4 balances expressiveness and numerical stability, while an order 4 introduces steeper curvature and modest instability.

\smallskip\noindent\textit{KAN Spline (KAS) Layer.} \quad This layer uses learnable, locally-supported B-spline basis functions to form activation functions, as defined in Eq.~(\ref{Eq:spline}). Each activation function is expressed as a linear combination of these splines, enabling the layer to adaptively model fine-grained patterns in the data. This localized, data-driven transformation improves the model's expressiveness, making it especially effective in medical image segmentation tasks where capturing subtle local variations is important.

The choice of Bernstein polynomials and splines as basis functions in our proposed model is motivated by their complementary approximation properties. Bernstein polynomials are globally supported over the interval $[0,1]$, meaning that each basis function influences the entire input domain. This global support results in smooth and stable approximations, where any change to a single coefficient affects the output across the full range. Such behavior is well-suited for modeling broad trends and ensuring continuity. In contrast, spline basis functions are locally supported, with each function active only within a limited region defined by a set of knots. This local support enables fine-grained adjustments: modifying a coefficient only alters the function's shape in a specific subregion, without impacting the rest of the domain. As a result, splines provide greater flexibility for capturing localized structures or sharp transitions. By integrating both types of basis functions, our model balances global smoothness with local adaptability, allowing for expressive and controllable function representations.

\smallskip\noindent After the tokenized feature maps are processed by the KAB and KAS layers, they go through the depth-wise convolution (DwConv), which applies a single filter per embedding dimension of the tokenized feature map, capturing local spatial patterns. The residual connection in the KABS block adds the input to the transformed output, preserving critical features. Layer normalization (LN) standardizes the embedding dimensions. The process of the KABS stage can be expressed as follows:
\begin{equation}
\small{\bm{Z}^{(k)} = \text{LN}(\text{DwConv}(\text{KAS}(\text{KAB}(\text{SE}(\bm{Z}^{(k-1)}))))+ \bm{Z}^{(k-1)})},
\label{Eq:KABS}
\end{equation}
where $\bm{Z}^{(k)}$ is the feature map of the $k$th KABS block for $k\in\{1,\dots,K\}$, and $\bm{Z}^{(0)}$ is the tokenized feature map of the last ConvSE block.

\subsection{Decoder}
The U-KABS decoder is designed as a symmetric counterpart to the encoder, processing features from the encoder's final block to produce a segmentation mask $\bm{M}\in [C]^{H\times W}$, where $C$ is the number of semantic classes in the input image. The mask generation process is carefully structured to gradually restore spatial resolution and refine features.

The decoder is composed of two primary stages, namely KABS and ConvSE, mirroring the encoder's design for architectural consistency. In the KABS stage, each of the two blocks begins by applying bilinear interpolation to upsample the input feature maps by a factor of two, ensuring a smooth and continuous increase in spatial resolution. The upsampled features are then concatenated with skip-connected features from the corresponding encoder layers to preserve fine-grained spatial information that may have been lost during downsampling. This skip connection strategy helps mitigate the semantic gap between encoder and decoder representations, enhancing localization accuracy. The subsequent ConvSE stage comprises three sequential blocks, each performing additional upsampling through bilinear interpolation and integrating encoder features via skip connections. These ConvSE blocks further refine the representations by emphasizing salient spatial and channel-wise patterns. To generate the segmentation mask, the output of the last ConvSE block is passed through a $1\times 1$ convolutional layer with $C$ output channels, mapping the features to a segmentation map. A softmax activation is applied along the channel dimension to compute class-wise probabilities for each pixel. The final segmentation mask $\bm{M}$ is obtained by taking the argmax over these probabilities, assigning each pixel the class with the highest probability.

\section{Experiments}

\subsection{Experimental Setup}
\noindent\textbf{Datasets.}\quad We conduct comprehensive experiments on four benchmark datasets: Breast Ultrasound Images (BUSI)~\cite{AlDhabyani2020}, Gland Segmentation (GlaS)~\cite{gland}, International Skin Imaging Collaboration (ISIC 2018)~\cite{codella2018skin}, and Automated Cardiac Diagnosis Challenge (ACDC)~\cite{bernard2018deep}.
\begin{itemize}
  \item \textbf{BUSI} includes ultrasound images of breast cancer, categorized as normal, benign, or malignant, with ground truth segmentation masks. We use 647 benign and malignant images, resized to $256\times256$.
  \item \textbf{GlaS} contains 165 histological colorectal tissue images, benign or malignant, with segmentation masks. Variability in staining and structural complexity challenges gland segmentation. Images are resized to $512\times512$.
  \item \textbf{ISIC 2018} comprises 2594 dermoscopic skin lesion images with binary segmentation masks. Blurry boundaries and diverse lesion types (benign, nevi, seborrheic keratoses, melanomas) pose challenges. Images are resized to $256\times256$.
  \item \textbf{ACDC} includes cine MRI heart scans with 5–8 mm slice thickness and 0.83–1.75 mm²/pixel resolution. Scans are annotated for left ventricle, right ventricle, myocardium, and background (four classes). It has 80 training and 20 validation cases.
\end{itemize}

\medskip\noindent\textbf{Evaluation Metrics.}\quad To compare the performance of U-KABS against the state-of-the-art (SOTA) methods, we adopt three standard evaluation metrics: Intersection Over Union (IoU), Dice Similarity Coefficient (DSC), and Hausdorff Distance (HD95). IoU measures the overlap between the predicted ($P$) and ground truth ($G$) segmentation masks:
\begin{equation}
\text{IoU} = \frac{|P\cap G|}{|P\cup G|},
\end{equation}
and DSC quantifies how well the predicted mask matches the ground truth mask:
\begin{equation}
DSC = \frac{2\times |P\cap G|}{|P| + |G|},
 \end{equation}
while HD95 measures the maximum deviation between the two masks, focusing on the 95th percentile of the nearest point distances:
\begin{equation}
\text{HD95}(P,G) = \max\{d_{95}(P,G), d_{95}(G,P)\}.
\end{equation}
Higher (resp. smaller) values of IoU and DSC (resp. HD95) indicate better segmentation performance.

\medskip\noindent\textbf{Baselines.}\quad We benchmark the performance of U-KABS against several SOTA methods, including U-Net~\cite{Ronneberger2015}, U-Net++~\cite{Zhou2018}, Attention-UNet~\cite{Oktay2018}, MedT~\cite{Valanarasu2021}, UNeXt~\cite{Valanarasu2022}, Rolling-Unet~\cite{Liu2024}, U-KAN~\cite{li2024ukan}, TPFIANet~\cite{zhao2025tpfianet}, ResU-KAN~\cite{wang2025resu}, MALUNet~\cite{Ruan2022}, ViT-CUP~\cite{Dosovitskiy2020}, Swin-Unet~\cite{Cao2021}, MCFormer~\cite{Li2023MCRformer}, FCT~\cite{Tragakis2023FCT}, UNETR~\cite{unetr}, TransUNet~\cite{Chen2024TransUNet}, SCM-UNET~\cite{yan2025scm} , WTCM-UNet, and CTO~\cite{lin2025rethinking} .

\medskip\noindent\textbf{Implementation Details.}\quad All experiments are conducted using PyTorch on a Linux workstation equipped with a single NVIDIA GeForce RTX 3070 GPU (8GB memory). For binary segmentation datasets, we use a batch size of 8 and a learning rate of $10^{-4}$, while for the multi-class ACDC dataset, the batch size is set to 2 with the same learning rate. The number of channels are set to $C_1=32$, $C_2=64$, $C_3=256$, $D_1=320$, and $D_2=512$. We set the Bernstein polynomial order to 4, spline order to 3, and grid size to 5. The model is trained using the Adam optimizer with a cosine annealing learning rate scheduler, decaying to a minimum of $10^{-5}$. We use patches of size $3\times 3$ (stride 2). For depth-wise convolution, tokens are reshaped to a 2D grid, a per-channel 2D depth-wise convolution (kernel $3\times3$) is applied across spatial dimensions, and the result is flattened back to tokens. For skip connections, the ConvSE blocks operate in image/grid space; their outputs are stored as skip features. After the KABS stage (tokens), features are untokenized (inverse patch projection) when needed for concatenation in the decoder. At each upsampling step of the decoder stage, we bilinearly upsample, untokenize (if needed), and concatenate with the corresponding encoder feature map before the subsequent ConvSE/KABS operations. Following prior work, we adopt the same standard evaluation protocols, including data split. Each dataset is randomly split into 80\% training and 20\% validation. For binary segmentation tasks, we employ the average of Binary Cross-Entropy (BCE) and Dice loss, while on the multi-class ACDC dataset, we use the average of Cross-Entropy and Dice loss. For all datasets, the model is trained for 400 epochs, and to ensure robustness, we report the mean and standard deviation over 5 independent runs with different random seeds.

\begin{table*}[!htb]
\caption{Quantitative comparison of our method against benchmark baselines on the BUSI and GlaS datasets in terms of IoU, DSC, and HD95. Best results are in bold, and the second best results are underlined. Entries marked with ``--'' indicate unavailable baseline code.}
\label{tab1}
\smallskip
\setlength\tabcolsep{6pt}
\centering
\small
\begin{tabular}{@{}lcccccccccc@{}}
 \toprule
 &  & & \multicolumn{3}{c}{BUSI} & \multicolumn{3}{c}{GlaS} \\
 \cmidrule(lr){4-6} \cmidrule(lr){7-9}
 Method  & Params(M) $\downarrow$& FLOPS(G)$\downarrow$ & IoU$\uparrow$ & DSC$\uparrow$ & HD95$\downarrow$ & IoU$\uparrow$ & DSC$\uparrow$ & HD95$\downarrow$ \\
 \midrule
 U-Net~\cite{Ronneberger2015} & 7.76 & 13.78 & 57.22$\pm$4.74 & 71.91$\pm$3.54 & 7.57$\pm$2.44 & 86.66$\pm$0.91 & 92.79$\pm$0.56 & 0.83$\pm$0.18 \\
 U-Net++~\cite{Zhou2018} & 9.16 & 34.90 & 57.41$\pm$4.77 & 72.11$\pm$3.90 & 7.72$\pm$2.16 & 87.07$\pm$0.76 & 92.96$\pm$0.65 & \textbf{0.81$\pm$0.16} \\
 Att-UNet~\cite{Oktay2018} & 50.27 & 60.06 & 55.18$\pm$3.61 & 70.22$\pm$2.88 & 8.36$\pm$2.11 & 86.84$\pm$1.19 & 92.89$\pm$0.65 & 0.82$\pm$0.29 \\
 MedT~\cite{Valanarasu2021} & \underline{1.60} & 21.24 & 52.15$\pm$3.47 & 67.68$\pm$3.18 & 10.23$\pm$1.17 & 75.47$\pm$3.46 & 85.92$\pm$2.93 & 1.04$\pm$0.12\\
 UNeXt~\cite{Valanarasu2022} & \textbf{1.47} & \textbf{4.58} & 61.78$\pm$1.46 & 75.52$\pm$0.91 & 8.33$\pm$0.42 & 83.95$\pm$1.09 & 91.22$\pm$0.67 & 1.04$\pm$0.10 \\
 Rolling-Unet~\cite{Liu2024} & 25.32 & 28.32 & 61.00$\pm$0.64 & 74.67$\pm$1.24 & \underline{6.19$\pm$0.62} & 86.42$\pm$0.96 & 92.63$\pm$0.62 & 1.00$\pm$0.08 \\
 U-KAN~\cite{li2024ukan} & 9.38 & \underline{6.89} & 63.38$\pm$2.83& 76.42$\pm$2.90& 7.19$\pm$0.18 & 87.64$\pm$0.32 & 93.37$\pm$0.16& 0.98$\pm$0.22 \\
 TPFIANet~\cite{zhao2025tpfianet} & -- & -- & 66.20$\pm$0.28& 75.84$\pm$0.26 & -- & 84.88$\pm$0.37 & 91.32$\pm$0.25 & -- \\
 ResU-KAN~\cite{wang2025resu} & 20.06 & 8.57 & \underline{67.74$\pm$1.35} & \underline{79.92$\pm$0.73} & 10.74$\pm$0.76 & \underline{87.99$\pm$0.44} & \underline{93.61$\pm$0.24 } & 1.01$\pm$ 0.06\\
 \midrule
 U-KABS (Ours) & 9.54 & 6.92 & \textbf{67.98$\pm$0.73} & \textbf{80.37$\pm$0.56} & \textbf{5.67$\pm$1.04} & \textbf{88.59$\pm$0.36} & \textbf{93.94$\pm$0.18} & \underline{0.82$\pm$0.09} \\
 \bottomrule
\end{tabular}
\end{table*}

\subsection{Results and Analysis}
\noindent\textbf{Quantitative Results.} \quad Table~\ref{tab1} compares our U-KABS model with SOTA methods on the BUSI and GlaS datasets, assessing model size (parameters), computational complexity (FLOPS, in gigaflops), and segmentation performance.  The reported results for the baselines are taken from their respective papers. U-KABS has a moderate parameter count, and achieves notably lower FLOPS, operating at 6.92 FLOPS(G), which is more efficient compared to the strongest baseline ResU-KAN. Our model demonstrates the most effective segmentation performance across both datasets. On the BUSI dataset, U-KABS achieves better performance than ResU-KAN in terms of both IoU and DSC. Similarly, on the GlaS dataset, U-KABS outperforms ResU-KAN in terms of both IoU and DSC . These improvements stem from U-KABS's hybrid architecture, enabling superior global and local feature approximation compared to KAN's B-spline-only approach. In addition, the integration of squeeze-and-excitation mechanisms into the convolutional blocks helps enhance feature prioritization in ultrasound (BUSI) and histological (GlaS) images.

\smallskip\noindent Table~\ref{tab:isic} shows that our model achieves the best performance compared to baselines in terms of both IoU and DSC on the ISIC 2018 dataset. U-KABS maintains a consistent performance advantage, with relative improvements of approximately 3.22\% in IoU and 1.9\% in DSC over the lightweight UNeXt. Similarly, U-KABS outperforms the strongest baseline, Rolling-Unet, achieving a relative improvement 0.55\% in terms of DSC. These improvements, though modest, are significant given the challenging nature of dermoscopic images in the ISIC 2018 dataset, which feature blurry boundaries and diverse lesion types. U-KABS's superior performance is largely attributed to its hybrid approach combining Bernstein polynomials and B-splines, enabling it to capture both global contextual trends and local fine-grained details more effectively than Rolling-Unet's blocks, which focus primarily on lightweight MLP-based feature learning. Moreover, the incorporation of squeeze-and-excitation mechanisms in U-KABS enhances its ability to prioritize salient features in complex skin lesion images.

\begin{table}[!htb]
\caption{Quantitative comparison of our method with benchmark baselines on ISIC 2018. Entries marked with ``--'' indicate unavailable baseline code.}
\label{tab:isic}
\smallskip
\setlength\tabcolsep{8pt}
\centering
\small
\begin{tabular}{@{}lccc@{}}
\toprule
Method & IoU$\uparrow$& DSC$\uparrow$ & HD95$\downarrow$\\
\midrule
U-Net~\cite{Ronneberger2015} & 77.86$\pm$ 0.96 & 87.55$\pm$ 0.87 & 1.79$\pm$ 0.23\\
U-Net++~\cite{Zhou2018} & 78.31 $\pm$ 0.65& 87.83$\pm$ 0.71 & \underline{1.56$\pm$ 0.09}\\
Att-UNet~\cite{Oktay2018} & 78.43$\pm$ 0.20 & 87.91$\pm$ 0.22 & 1.69$\pm$ 0.14\\
MedT~\cite{Valanarasu2021} & 81.48 $\pm$ 0.26& 89.49$\pm$ 0.18 & 1.89$\pm$ 0.31\\
UNeXt~\cite{Valanarasu2022} & 81.70 $\pm$ 1.53 & 89.70$\pm$ 0.96 & 1.66$\pm$ 0.18\\
MALUNet~\cite{Ruan2022} & 80.25$\pm$ 0.56 & 89.04$\pm$ 0.32 & 2.73$\pm$ 0.10 \\
Rolling-Unet~\cite{Liu2024} & 83.74$\pm$ 0.16 & \underline{90.90$\pm$ 0.12} & 1.99$\pm$ 0.07\\
SCM-UNET~\cite{yan2025scm} & 81.88 & 90.14 & --\\
WTCM-UNet~\cite{gan2026wtcm} & 80.26 & 89.05& --\\
CTO~\cite{lin2025rethinking} & \underline{84.00$\pm$ 0.29} & 90.60$\pm$ 0.17& 3.78$\pm$ 0.15 \\
\midrule
Ours & \textbf{84.33$\pm$ 0.06} & \textbf{91.40$\pm$ 0.04} & \textbf{1.48$\pm$ 0.07}\\
\bottomrule
\end{tabular}
\end{table}

\begin{table}[!htb]
\caption{Quantitative comparison of our model with benchmark baselines on the ACDC dataset in terms of DSC, highlighting per-organ performance for the right ventricle (RV), myocardium (Myo), and left ventricle (LV), along with the average DSC scores.}
\label{tab:acdc}
\smallskip
\setlength\tabcolsep{10pt}
\centering
\begin{tabular}{@{}lccccc@{}}
\toprule
 & \multicolumn{3}{c}{ACDC} &  \\
\cmidrule{2-4}
Method & RV & Myo & LV & Avg (\%)\\
\midrule
ViT-CUP~\cite{Dosovitskiy2020} & 81.46 & 70.71 & 92.18 & 81.45 \\
Swin-Unet~\cite{Cao2021} & 88.55 & 85.62 & \textbf{95.83} & \underline{90.00} \\
UNETR~\cite{unetr} & 85.29 & \underline{86.52} & 94.02 & 88.61 \\
MCFormer~\cite{Li2023MCRformer} & 87.60 & 79.20 & 93.40 & 87.70 \\
FCT~\cite{Tragakis2023FCT} & 86.10 & 81.90 & 94.80 & 87.60 \\
TransUNet~\cite{Chen2024TransUNet} & \underline{88.86} & 84.53 & \underline{95.73} & 89.71 \\
\midrule
Ours & \textbf{88.97} & \textbf{87.80} & 94.76 & \textbf{90.53} \\
\bottomrule
\end{tabular}
\end{table}

\begin{figure*}[!htb]
\centering
\includegraphics[width=\linewidth]{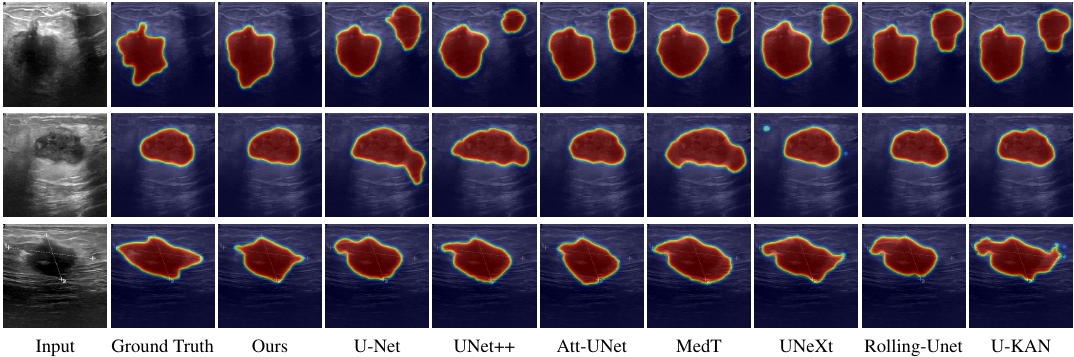}
\caption{Qualitative evaluation of our method using heatmaps compared to benchmark baselines on the BUSI dataset.}
\label{fig:busi}
\end{figure*}

\smallskip\noindent Table~\ref{tab:acdc} presents the performance comparison of U-KABS against baselines on the ACDC dataset, evaluated using the DSC metric for three cardiac structures (Right Ventricle (RV), Myocardium (Myo), and Left Ventricle (LV)), along with their average. U-KABS achieves a 0.53\% higher average DSC than Swin-Unet, which is the best-performing baseline. When compared to TransUNet, U-KABS improves by 0.82\% in average DSC, and the gain over UNETR is 1.92\%. The difference becomes even more significant when compared to ViT-CUP, with U-KABS outperforming it by a large margin of 9.08\% in average DSC. On a per-organ level, U-KABS shows a 2.18\% improvement over Swin-Unet for myocardium segmentation and a 0.42\% improvement for the right ventricle. Although Swin-Unet slightly outperforms U-KABS on the left ventricle, U-KABS shows a more balanced performance across all the three structures.

\begin{figure*}[!htb]
\centering
\includegraphics[width=\linewidth]{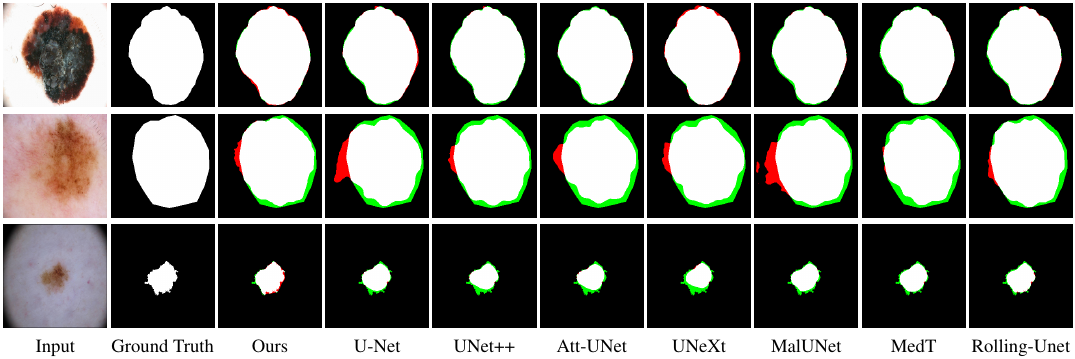}
\caption{Qualitative comparison of our method with benchmark baselines on the ISIC 2018 dataset, highlighting segmentation errors. White, red, and green regions represent predicted segmentation, over-segmentation, and under-segmentation, respectively}
\label{fig:isic_vis}
\end{figure*}

\smallskip\noindent\textbf{Qualitative Results.} \quad To assess the effectiveness of U-KABS qualitatively, we conduct visual comparisons of results on the BUSI, ISIC 2018, and ACDC datasets. Figure~\ref{fig:busi} presents a visual comparison of segmentation outputs generated by U-KABS and baselines on the BUSI dataset using heatmaps. The figure includes three representative cases, each displaying the input ultrasound image, the ground truth mask, and segmentation heatmaps. The ground truth mask serves as the reference for evaluating segmentation accuracy. U-KABS demonstrates superior alignment with the ground truth, accurately capturing lesion boundaries with minimal false positives or under-segmented regions, even in challenging cases with low-contrast or noisy ultrasound images. For instance, in the top row, U-KABS precisely delineates irregular lesion contours, whereas U-Net and UNet++ produce diffused heatmaps with over-segmentation around lesion edges, leading to higher false positives. Similarly, in the middle row, U-KABS captures fine-grained boundary details. In the bottom row, U-KABS outperforms U-KAN by reducing under-segmentation in heterogeneous lesion regions, attributed to the hybrid KABS block combining Bernstein polynomials for global smoothness and B-splines for local adaptability. Attention U-Net and Rolling-Unet show improved boundary focus but still exhibit artifacts, such as jagged edges, compared to U-KABS's smooth delineations. These qualitative results highlight U-KABS's robustness in handling ultrasound-specific challenges, such as low contrast, supporting its quantitative superiority (e.g., 80.37\% DSC on BUSI, Table~\ref{tab1}). The integration of squeeze-and-excitation mechanisms enhances feature prioritization, enabling U-KABS to emphasize salient lesion regions over background noise, as evidenced by the focused heatmaps.

\begin{figure*}[!htb]
\centering
\includegraphics[width=0.82\linewidth]{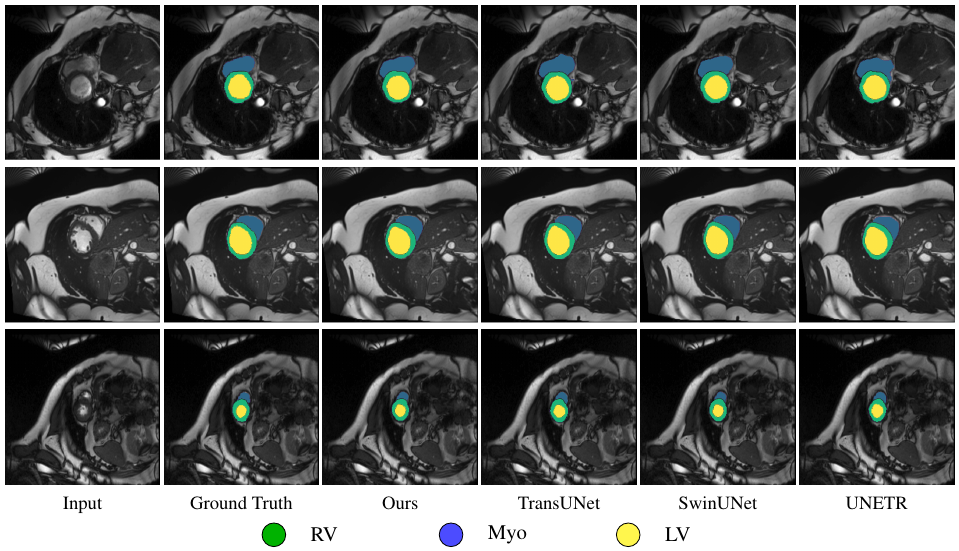}
\caption{Qualitative comparison of our model with baselines on the ACDC dataset, where RV, Myo, LV represent Right Ventricle, Myocardium and Left Ventricle, respectively.}
\label{fig:acdc_vis}
\end{figure*}

\smallskip\noindent Figure~\ref{fig:isic_vis} illustrates the qualitative segmentation performance of U-KABS on representative dermoscopic images from the ISIC 2018 skin lesion dataset, highlighting its ability to delineate lesion boundaries accurately. The figure includes the input dermoscopic image, the corresponding ground truth mask, and segmentation outputs for U-KABS and baselines. U-KABS most accurately matches the ground truth, outlining lesions with more precise boundaries. These qualitative results highlight U-KABS's robustness in addressing ISIC 2018's challenges, such as blurry boundaries, diverse lesion appearances, and imaging artifacts, reinforcing its quantitative superiority.

\smallskip\noindent Figure~\ref{fig:acdc_vis} presents qualitative comparisons of UKABS with benchmark baselines on the ACDC dataset. Owing to the intrinsic characteristics of cardiac MRI segmentation and the substantial anatomical overlap between the right ventricle (RV) and left ventricle (LV), visual distinctions among different approaches are inherently subtle, particularly for larger and anatomically contiguous structures such as the LV. Nevertheless, the qualitative results remain consistent with the quantitative evaluation, wherein UKABS demonstrates subtle yet discernible improvements over well-established baselines, including TransUNet, Swin-Unet, and UNETR.

\subsection{Model Efficiency Analysis}
\noindent Table~\ref{tab:latency_fps} reports a comparative evaluation of inference efficiency across different segmentation methods in terms of latency per image and frames per second (FPS). Among all evaluated approaches, UNeXt achieves the lowest latency (3.06\,ms) and the highest throughput (326.79 FPS), reflecting its computationally efficient design and suitability for point-of-care deployment. ResUKAN also demonstrates strong real-time performance, achieving a low latency of 5.57\,ms and a high FPS of 181.23, which can be attributed to the use of residual connections that facilitate effective information propagation with minimal additional computational overhead. U-KAN maintains a balanced trade-off between efficiency and segmentation accuracy, yielding a latency of 9.66\,ms and an FPS of 103.54. In contrast, the proposed U-KABS model exhibits higher latency (18.62\,ms) and lower FPS (53.71) compared to U-KAN and ResUKAN; however, it consistently outperforms baseline methods in terms of segmentation accuracy. Notably, U-KABS remains more efficient than ResUKAN with respect to FLOPs and model parameters, indicating a well-balanced compromise between training and inference efficiency and overall segmentation performance.

\begin{table}[!htb]
\centering
\caption{Latency and frames per second (FPS) comparison.}
\label{tab:latency_fps}
\smallskip
\centering
\begin{tabular}{@{}lcc@{}}
\toprule
Method & Latency per image (ms)$\downarrow$ & FPS$\uparrow$ \\
\midrule
MedT~\cite{Valanarasu2021}         & 305.71 & 3.27   \\
Rolling-Unet~\cite{Liu2024}           & 55.56  & 18.00  \\
UNeXt~\cite{Valanarasu2022}       & 3.06   & 326.79 \\
U-KAN~\cite{li2024ukan}               & 9.66   & 103.54 \\
ResUKAN~\cite{wang2025resu}   & 5.57   & 181.23 \\
\midrule
Ours         & 18.62  & 53.71  \\
\bottomrule
\end{tabular}
\end{table}

\begin{figure}[!ht]
\centering
\includegraphics[width=.97\linewidth]{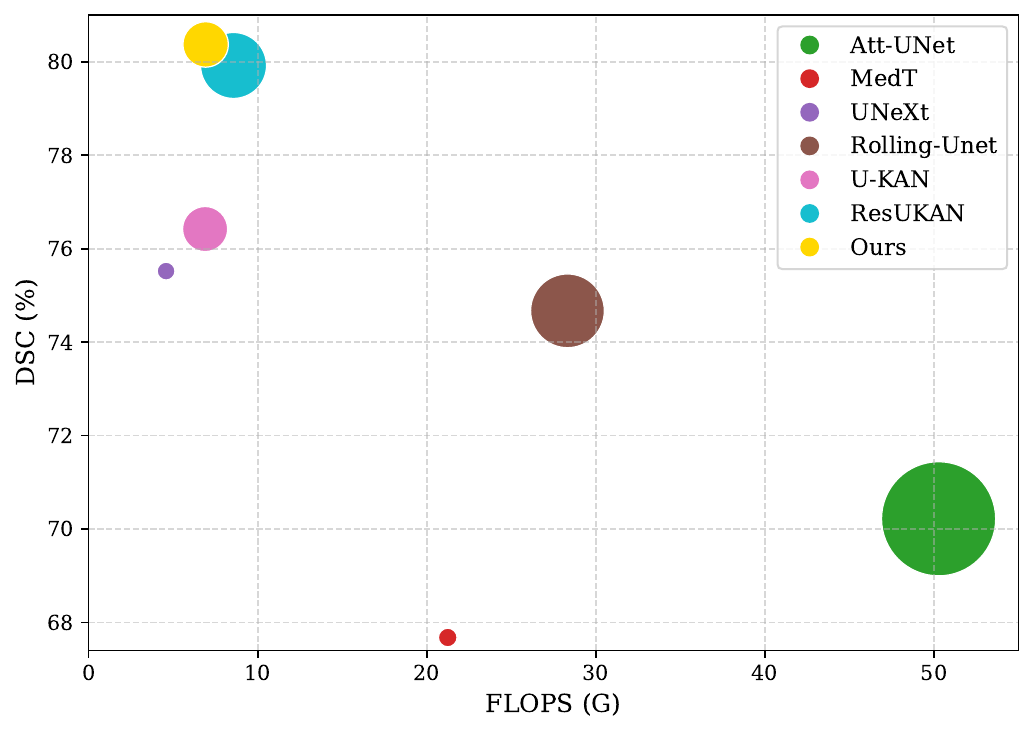}
\caption{Comparison of performance and model efficiency on the BUSI dataset. Our U-KABS model is benchmarked against state-of-the-art methods, including Attention-UNet~\cite{Oktay2018}, MedT~\cite{Valanarasu2021}, UNeXt~\cite{Valanarasu2022}, Rolling-Unet~\cite{Liu2024}, U-KAN~\cite{li2024ukan}, and ResU-KAN~\cite{wang2025resu}. Performance is assessed using the DSC metric, where higher values indicate superior segmentation performance. The size of the circle indicate the number of learnable parameters..}
\label{Fig:ModelEfficiency}
\end{figure}

\smallskip\noindent Figure~\ref{Fig:ModelEfficiency} presents a circle chart comparing the segmentation performance and model efficiency of U-KABS against state-of-the-art baselines on the BUSI dataset, evaluated using the DSC metric against floating-point operations per second (FLOPS, in gigaflops). The circle size represents the number of learnable parameters, reflecting model size. U-KABS achieves a superior DSC of 80.37\% with a moderate 9.54M parameters, striking a good balance between performance and efficiency, as visualized in the chart. Compared to lightweight models, UNeXt and MedT offer efficiency but exhibit under-segmentation due to limited expressiveness of their MLP-based architectures. U-KABS's moderate parameter count is justified by its performance gains, driven by the hybrid design of the KABS block, which integrates Bernstein polynomials for global smoothness and B-splines for local adaptability.

\subsection{Ablation Study}
Table~\ref{Tab:Ablation} assesses the impact of the core components of our network architecture on model performance using the BUSI dataset. We designate the baseline configuration as the model without the KAB layer. This baseline is marked in the first row of the table by an asterisk appended to the IoU and DSC scores. In each subsequent row, we remove exactly one component from the U-KABS model (indicated by a lightgray cross) to isolate its effect and quantify the corresponding change in performance. The complete model, incorporating all components, yields the best performance. Excluding the KAB layer leads to the most significant performance decline, with IoU and DSC scores dropping to 61.74\% and 75.98\%, respectively, underscoring its pivotal role as a global approximator. In contrast, excluding the KAS layer while retaining its KAB counterpart results in superior performance. Omitting other components reduces IoU by 0.91-6.24\% and DSC by 0.48-4.39\%, indicating their substantial contributions. Notably, the global approximation capability of the KAB layer, when combined with the local adaptability of other components, leads to a synergistic improvement in segmentation accuracy.

\begin{table}[!htb]
\caption{Effect of individual core components on model performance using the BUSI dataset, where ``*'' indicates the baseline.}
\label{Tab:Ablation}
\smallskip
\setlength\tabcolsep{5pt} 
\centering
\begin{tabular}{ccccccc}
\toprule
\multicolumn{5}{c}{Component} & \multicolumn{2}{c}{Metric (\%)} \\
\cmidrule(lr){1-5} \cmidrule(lr){6-7}
KAB & KAS & Tokenize & SE & DWConv & IoU & DSC\\
\midrule
\xmark & \cmark & \cmark & \cmark & \cmark & 61.74$^{\textcolor{blue}{*}}$ & 75.98$^{\textcolor{blue}{*}}$ \\
\cmark & \xmark & \cmark & \cmark & \cmark & 64.28 & 77.92 \\
\cmark & \cmark & \xmark & \cmark & \cmark & 62.48 & 76.25 \\
\cmark & \cmark & \cmark & \xmark & \cmark & 65.61 & 78.34 \\
\cmark & \cmark & \cmark & \cmark & \xmark & 67.07 & 79.89 \\
\cmark & \cmark & \cmark & \cmark & \cmark & \textbf{67.98} & \textbf{80.37} \\
\bottomrule
\end{tabular}
\end{table}

\noindent\textbf{Effect of Squeeze-and-Excitation Mechanism.} \quad Table~\ref{Tab:SE} presents the impact of incorporating squeeze-and-excitation (SE) mechanisms within the convolutional stage of the U-KABS framework, evaluated on the BUSI dataset using the IoU and DSC metrics. The table compares U-KABS with and without SE in its convolutional stages, demonstrating that the inclusion of SE improves both metrics. These gains highlight the SE mechanism's role in enhancing U-KABS's segmentation performance, particularly for challenging ultrasound images characterized by low contrast, and heterogeneous lesion textures.

\begin{table}[!htb]
\caption{Effect of squeeze-and-excitation (SE) on model performance using the BUSI dataset.}
\smallskip
\setlength\tabcolsep{10pt}
\centering
\begin{tabular}{@{}lcc@{}}
\toprule
Convolutional Stage & IoU & DSC \\
\midrule
Without SE & 66.91 & 79.73 \\
With SE & \textbf{67.98} & \textbf{80.37} \\
 \bottomrule
\end{tabular}
\label{Tab:SE}
\end{table}

\medskip\noindent\textbf{Effect of Bernstein Polynomial Order.} \quad Table~\ref{Tab:BernOrder} evaluates the impact of varying the Bernstein polynomial order in the KAB layer of U-KABS on segmentation performance, assessed on the BUSI dataset using the IoU and DSC metrics. The Bernstein polynomial order in the KAB layer defines the degree of the basis polynomials, which govern the global smoothness of feature approximations in U-KABS. Higher orders increase the polynomial's capacity to model complex functions but risk overfitting or numerical instability due to increased parameters. The table shows that order 4 achieves the best performance, indicating a good balance between expressiveness and stability for ultrasound image segmentation thanks, in large part, to its ability to capture smooth, continuous lesion shapes in BUSI images. This polynomial order effectively models global anatomical trends, such as heterogeneous lesion textures, without over-smoothing or underfitting, unlike order 3, which underfits due to limited flexibility, resulting in under-segmentation. Order 5's slight performance drop suggests overfitting or excessive smoothness, reducing precision for fine-grained boundaries, critical for ultrasound images with low contrast.

\begin{table}[!htb]
\caption{Effect of Bernstein polynomial order on model performance using the BUSI dataset.}
\smallskip
\setlength\tabcolsep{10pt}
\centering
\begin{tabular}{@{}ccc@{}}
\toprule
Bernstein Polynomial Order &  IoU & DSC\\
\midrule
2 & 65.92 & 79.08\\
3 & 65.80 & 78.99 \\
4 & \textbf{67.98} & \textbf{80.37}\\
5 & 65.34 & 78.67 \\
\bottomrule
\end{tabular}
\label{Tab:BernOrder}
\end{table}

\medskip\noindent\textbf{Effect of Spline Order and Grid Size.} \quad Table~\ref{Tab:SplineOrderGrid} investigates the impact of varying the B-spline order and grid size in the KAS layer of the U-KABS framework on segmentation performance, evaluated on the BUSI dataset using the IoU and DSC metrics. The spline order controls the polynomial degree, affecting smoothness and flexibility, while grid size governs the granularity of local approximations, with larger grid sizes increasing the number of basis functions for finer control. The table shows that the combination of spline order 3 and grid size 5 achieves the best performance, demonstrating its effectiveness in capturing fine-grained lesion boundaries in BUSI images. The KAS layer's B-splines complement the KAB layer's Bernstein polynomials, which provide global smoothness, by refining local details critical for ultrasound challenges low contrast.

\begin{table}[!htb]
\caption{Effect of spline order and grid size on model performance using the BUSI dataset.}
\smallskip
\setlength\tabcolsep{10pt}
\centering
\begin{tabular}{@{}cccc@{}}
\toprule
Spline Order & Grid Size & IoU & DSC \\
\midrule
 & 3 & 66.00 & 79.17 \\
 2 & 4 & 66.41 & 79.40 \\
 & 5 & 65.15 & 78.74 \\
\midrule
& 3 & 65.98 & 79.07 \\
3 & 4 & 67.19 & 80.10 \\
& 5 & \textbf{67.98} & \textbf{80.37} \\
\midrule
& 3 & 64.62 & 78.13 \\
4 & 4 & 67.42 & 79.93 \\
& 5 & 66.97 & 79.61 \\
\bottomrule
\end{tabular}
\label{Tab:SplineOrderGrid}
\end{table}

\medskip\noindent\textbf{Comparison with Standard Transformer and Residual Convolution Blocks.}\quad Table~\ref{Tab:Blocks} presents a quantitative comparison between the proposed KABS block and alternative architectural components, namely a residual convolution block and a standard Transformer block, within the same network backbone on the BUSI dataset. Replacing the KABS block with a residual convolutional block leads to a notable degradation in segmentation performance, with reductions of 6.56\% in IoU and 5.48\% in DSC. Similarly, although the standard Transformer block, comprising multi-head self-attention and an MLP, outperforms the residual convolutional block, it still exhibits an observable performance gap when compared to the KABS block. These results highlight the effectiveness of the proposed KABS design in capturing discriminative features and enhancing segmentation quality beyond what is achievable using conventional Transformer or residual convolutional blocks alone.

\begin{table}[!htb]
\caption{Performance comparison of proposed KABS block with standard Transformer and residual convolution blocks.}
\label{Tab:Blocks}
\setlength\tabcolsep{15pt}
\smallskip
\centering
\begin{tabular}{@{}lcc@{}}
\toprule
Method & IoU & DSC \\
\midrule
Residual Convolution & 61.42 & 74.89 \\
Standard Transformer & 63.71 & 76.45 \\
KABS (ours) & \textbf{67.98} & \textbf{80.37} \\
 \bottomrule
\end{tabular}
\end{table}

\section{Discussion}
As demonstrated in the experiments, U-KABS achieves strong performance across diverse datasets while maintaining computational efficiency. The results highlight its strengths in robust global-local feature modeling, enabled by the hybrid KAN Bernstein spline block, which combines Bernstein polynomials for smooth global approximations and B-splines for precise local refinements. This design addresses key challenges in medical imaging, such as noise in ultrasound and blurry boundaries in dermoscopy, as evidenced by the qualitative results, where U-KABS exhibits minimal over- or under-segmentation compared to baselines like U-Net and TransUNet. The ablation studies further underscore these strengths: the KAB layer's global approximation is pivotal, while the KAS layer's local adaptability and squeeze-and-excitation mechanisms enhance feature prioritization, reducing artifacts in noisy regions. Unlike Transformer-based models, U-KABS achieves higher accuracy with substantially lower computational cost. Despite these strengths, U-KABS has two main limitations that warrant consideration. First, its parameter count is higher than lightweight baselines such as UNeXt, potentially limiting deployment on edge devices with memory constraints, although its FLOPS remains competitive. Second, while ablation studies confirm component contributions, broader generalization across unseen modalities (e.g., PET) requires additional validation. 

\section{Conclusion}
We introduced U-KABS, a novel hybrid framework that integrates KANs with a U-shaped encoder-decoder architecture, leveraging the global smoothness of Bernstein polynomials and the local adaptability of B-splines to effectively capture both broad contextual trends and fine-grained patterns in medical images. Extensive experiments on various benchmark datasets demonstrate that U-KABS consistently outperforms state-of-the-art methods. Notably, U-KABS excels in segmenting complex anatomical structures and performs robustly in both binary and multi-class segmentation tasks. Qualitative results further demonstrate U-KABS's ability to produce accurate segmentation masks, with improved boundary delineation. Moreover, ablation results confirm the importance of the core architectural components, especially the KAB layer for global approximation. For future work, we aim to further optimize U-KABS for efficiency, and incorporate self-supervised learning to improve performance on datasets with limited annotated data, a common challenge in medical imaging.

\subsection*{Statements and Declarations}
\noindent{\textbf{Author Contribution Declaration}}\quad D.B. conducted the comprehensive experiments, including model training and evaluation on benchmark datasets with ablation studies, and wrote the initial draft focusing on methodology and results. A.Y. and A.B.H. conceptualized and designed the proposed  framework, and provided overall supervision and critical revisions to the final draft for clarity, novelty, and technical rigor. All authors contributed to the literature review, qualitative analysis, and efficiency evaluation, reviewed the manuscript for scientific accuracy, and approved the final version.

\smallskip\noindent{\textbf{Conflict of Interest}}\quad The authors declare that they have no financial or personal interests to disclose.

\smallskip\noindent{\textbf{Funding}}\quad This work was supported in part by Natural Sciences and Engineering Research Council of Canada.

\smallskip\noindent{\textbf{Data Availability}}\quad The datasets used in the experiments are publicly available

\bibliographystyle{ieeetr}
\bibliography{references}

@article{AlDhabyani2020,
  author = {Al-Dhabyani, Wael and Gomaa, Mohammed and Khaled, Hussien and Fahmy, Aly},
  title = {Dataset of breast ultrasound images},
  journal = {Data in Brief},
  volume = {28},
  pages = {},
  year = {2020}
}

@inproceedings{Ronneberger2015,
  author = {Ronneberger, Olaf and Fischer, Philipp and Brox, Thomas},
  title = {{U-Net: Convolutional networks for biomedical image segmentation}},
  booktitle = {Proc. International Conference on Medical Image Computing and Computer-Assisted Intervention},
  pages = {234--241},
  year = {2015},
}

@inproceedings{Zhou2018,
  author = {Zhou, Zongwei and Siddiquee, Md Mahfuzur Rahman and Tajbakhsh, Nima and Liang, Jianming},
  title = {{UNet++: A nested U-Net architecture for medical image segmentation}},
  booktitle = {Deep Learning in Medical Image Analysis and Multimodal Learning for Clinical Decision Support},
  pages = {3--11},
  year = {2018}
}

@inproceedings{Valanarasu2022,
  author = {Valanarasu, Jeya Maria Jose and Patel, Vishal M.},
  title = {{UNeXt: MLP-based rapid medical image segmentation network}},
  booktitle = {Proc. International Conference on Medical Image Computing and Computer-Assisted Intervention},
  pages = {23--33},
  year = {2022},
}

@inproceedings{Oktay2018,
  author = {Oktay, Ozan and Schlemper, Jo and Le Folgoc, Loic and Lee, Michael and Heinrich, Mattias and Misawa, Kazunari and Mori, Kensaku and McDonagh, Steven and Hammerla, Nils Y. and Kainz, Bernhard and Glocker, Bernhard and Rueckert, Daniel},
  title = {{Attention U-Net: Learning where to look for the pancreas}},
  booktitle = {Proc. Medical Imaging with Deep Learning},
  year = {2018}
}

@inproceedings{Cicek2016,
  author = {Ozgun Cicek and Ahmed Abdulkadir and Soeren S. Lienkamp and Thomas Brox and Olaf Ronneberger},
  title = {{3D U-Net: Learning dense volumetric segmentation from sparse annotation}},
  booktitle = {Proc. International Conference on Medical Image Computing and Computer-Assisted Intervention},
  pages = {424--432},
  year = {2016}
}

@inproceedings{Huang2020,
  author = {Huang, Huimin and Lin, Lanfen and Tong, Rui and Hu, Hongjie and Zhang, Qiaowei and Iwamoto, Yutaro and Han, Xianhua and Chen, Yen-Wei and Wu, Jian},
  title = {{UNet 3+: A full-scale connected UNet for medical image segmentation}},
  booktitle = {Proc. IEEE International Conference on Acoustics, Speech and Signal Processing},
  year = {2020}
}

@inproceedings{Milletari2016,
  author = {Milletari, Fausto and Navab, Nassir and Ahmadi, Seyed-Ahmad},
  title = {{V-Net: Fully convolutional neural networks for volumetric medical image segmentation}},
  booktitle = {Proc. International Conference on 3D Vision},
  pages = {565--571},
  year = {2016}
}

@article{Liu2024,
  author = {Liu, Yutong and Zhu, Haijiang and Liu, Mengting and Yu, Huaiyuan and Chen, Zihan and Gao, Jie},
  title = {{Rolling-Unet: Revitalizing MLP's ability to efficiently extract long-distance dependencies for medical image segmentation}},
  journal = {Proc. AAAI Conference on Artificial Intelligence},
  year = {2024}
}

@article{Chen2024TransUNet,
  title={{TransUNet: Rethinking the U-Net architecture design for medical image segmentation through the lens of transformers}},
  author={Chen, Jieneng and Mei, Jieru and Li, Xianhang and Lu, Yongyi and Yu, Qihang and Wei, Qingyue and Luo, Xiangde and Xie, Yutong and Adeli, Ehsan and Wang, Yan and others},
  journal={Medical Image Analysis},
  volume={97},
  pages={},
  year={2024}
}

@inproceedings{Valanarasu2021,
  author = {Valanarasu, Jeya Maria Jose and Oza, Poojan and Hacihaliloglu, Ilker and Patel, Vishal M.},
  title = {{Medical Transformer: Gated axial-attention for medical image segmentation}},
  booktitle = {Proc. International Conference on Medical Image Computing and Computer-Assisted Intervention},
  pages = {36--46},
  year = {2021}
}

@inproceedings{Ruan2022,
  title = {{MALUNet: A multi-attention and light-weight UNet for skin lesion segmentation}},
  author = {Ruan, Jiacheng and Xiang, Suncheng and Xie, Mingye and Liu, Ting and Fu, Yuzhuo},
  booktitle = {Proc. IEEE International Conference on Bioinformatics and Biomedicine},
  year = {2022}
}

@inproceedings{Cao2021,
  title={{Swin-Unet: Unet-like pure transformer for medical image segmentation}},
  author={Cao, Hu and Wang, Yueyue and Chen, Joy and Jiang, Dongsheng and Zhang, Xiaopeng and Tian, Qi and Wang, Manning},
  booktitle={Proc. European Conference on Computer Vision},
  pages={205--218},
  year={2022}
}

@inproceedings{Liu2024KAN,
  author = {Liu, Ziming and Wang, Yixuan and Vaidya, Sachin and Ruehle, Fabian and Halverson, James and Soljačić, Marin and Hou, Thomas Y. and Tegmark, Max},
  title = {{KAN: Kolmogorov-Arnold networks}},
  booktitle = {International Conference on Learning Representations},
  year = {2025}
}

@inproceedings{Dosovitskiy2020,
  author = {Dosovitskiy, Alexey and Beyer, Lucas and Kolesnikov, Alexander and Weissenborn, Dirk and Zhai, Xiaohua and Unterthiner, Thomas and Dehghani, Matthias and Minderer, Matthias and Heigold, Georg and Gelly, Sylvain and Uszkoreit, Jakob and Houlsby, Neil},
  title = {{An image is worth 16x16 words: Transformers for image recognition at scale}},
  booktitle = { International Conference on Learning Representations},
  year = {2021},
}

@inproceedings{Valanarasu2020a,
  author = {Valanarasu, Jeya Maria Jose and Sindagi, Vishwanath A. and Hacihaliloglu, Ilker and Patel, Vishal M.},
  title = {{Towards accurate segmentation of biomedical images using over-complete representations}},
  booktitle = {{Proc. International Conference on Medical Image Computing and Computer-Assisted Intervention }},
  pages = {363--373},
  year = {2020},
}

@article{Valanarasu2020b,
  author = {Valanarasu, Jeya Maria Jose and Yasarla, Rajeev and Wang, Puyang and Hacihaliloglu, Ilker and Patel, Vishal M.},
  title = {{Learning to segment brain anatomy from 2D ultrasound with less data}},
  journal = {IEEE Journal of Selected Topics in Signal Processing},
  volume = {14},
  number = {6},
  pages = {1221--1234},
  year = {2020}
}

@inproceedings{li2024ukan,
  title = {{U-KAN makes strong backbone for medical image segmentation and generation}},
  author = {Li, Chenxin and Liu, Xinyu and Li, Wuyang and Wang, Cheng and Liu, Hengyu and Yuan, Yixuan},
 booktitle = {Proc. AAAI Conference on Artificial Intelligence},
  year = {2025},
}

@inproceedings{dwconv,
    title = {{Xception: Deep learning with depthwise
separable convolutions}},
author = {François Chollet},
    booktitle = {Proc. IEEE Conference on Computer Vision and Pattern Recognition},
    year = {2017},
}

@inproceedings{unetr,
  title={{UNETR: Transformers for 3D medical image segmentation}},
  author={Hatamizadeh, Ali and Tang, Yucheng and Nath, Vishwesh and Yang, Dong and Myronenko, Andriy and Landman, Bennett and Roth, Holger R and Xu, Daguang},
  booktitle={Proc. IEEE Winter Conference on Applications of Computer Vision},
  pages={574--584},
  year={2022}
}

@article{Braun2009KANTH,
  title={On a constructive proof of Kolmogorov’s superposition theorem},
  author={Braun, J{\"u}rgen and Griebel, Michael},
  journal={Constructive approximation},
  volume={30},
  pages={653--675},
  year={2009},
}

@article{Johannes2021KANTH,
  title={The {Kolmogorov-Arnold} representation theorem revisited},
  author={Schmidt-Hieber, Johannes},
  journal={Neural networks},
  volume={137},
  pages={119--126},
  year={2021},
}

@article{bernard2018deep,
  title={{Deep learning techniques for automatic MRI cardiac multi-structures segmentation and diagnosis: Is the problem solved?}},
  author={Bernard, Olivier and Lalande, Alain and Zotti, Clement and Cervenansky, Frederick and Yang, Xin and Heng, Pheng-Ann and Cetin, Irem and Lekadir, Karim and Camara, Oscar and Ballester, Miguel Angel Gonzalez},
  journal={IEEE Transactions on Medical Imaging},
  volume={37},
  number={11},
  pages={2514--2525},
  year={2018}}

@inproceedings{codella2018skin,
  title={Skin lesion analysis toward melanoma detection: A challenge at the 2017 International Symposium on Biomedical Imaging},
  author={Codella, Noel CF and Gutman, David and Celebi, M Emre and Helba, Brian and Marchetti, Michael A and Dusza, Stephen W and Kalloo, Aadi and Liopyris, Konstantinos and Mishra, Nabin and Kittler, Harald},
  booktitle={Proc. IEEE International Symposium Conference on Biomedical Imaging},
  year={2018}}

@article{gland,
  title={Gland segmentation in colon histology images: The glas challenge contest},
  author={Sirinukunwattana, Korsuk and Pluim, Josien PW and Chen, Hao and Qi, Xiaojuan and Heng, Pheng-Ann and Guo, Yun Bo and Wang, Li Yang and Matuszewski, Bogdan J and Bruni, Elia and Sanchez, Urko and others},
  journal={Medical Image Analysis},
  volume={35},
  pages={489--502},
  year={2017}
}

@article{lin2025rethinking,
	title={{Rethinking boundary detection in deep learning-based medical image segmentation}},
	author={Lin, Yi and Zhang, Dong and Fang, Xiao and Chen, Yufan and Cheng, Kwang-Ting and Chen, Hao},
	journal={Medical Image Analysis},
	pages={},
	year={2025}
}

@article{zhao2025tpfianet,
	title={{TPFIANet: Three path feature progressive interactive attention learning network for medical image segmentation}},
	author={Zhao, Yawu and Wang, Shudong and Ren, Yande and Wang, Jiehuan and Wang, Shaoqiang and Qiao, Sibo and Liu, Tiyao and Pang, Shanchen},
	journal={Knowledge-Based Systems},
	pages={},
	year={2025}
}

@article{wang2025resu,
	title={{ResU-KAN}: a medical image segmentation model integrating residual convolutional attention and atrous spatial pyramid pooling},
	author={Wang, Haibin and Zhao, Zhenfeng and Liu, Qi and Wang, Shenwen},
	journal={Applied Intelligence},
	volume={55},
	number={7},
	pages={},
	year={2025}
}

@article{gan2026wtcm,
	title={{WTCM-UNet:} A hybrid {CNN-SSM} framework combining wavelet transform for medical image segmentation},
	author={Gan, Zhihua and Xie, Zhongxiang and Zhang, Yushu and Han, Weihong and Zhang, Bo and Chai, Xiuli},
	journal={Biomedical Signal Processing and Control},
	volume={112},
	pages={},
	year={2025}
}

@article{yan2025scm,
	title={{SCM-UNet}: Spatial-Channel Mamba {UNet} for medical image segmentation},
	author={Yan, Haijie and Hong, Qiuhong and Wei, Shoulin and Zhang, Xiangliang and Yin, Jibin},
	journal={Digital Signal Processing},
	pages={},
	year={2025}
}

@article{Li2023MCRformer,
	title={{MCRformer}: Morphological Constraint Reticular Transformer for {3D} Medical Image Segmentation},
	author={Jun Li  and Nan Chen and Han Zhou and Riqing Chen  and Chunhui Feng  and Heng Dong and Taotao Lai and Changcai Yang and Lifang Wei and Fanggang Cai},
	journal={Expert Systems with Applications},
	volume={232},
	pages={},
	year={2023}
}

@inproceedings{Tragakis2023FCT,
	title={The Fully Convolutional Transformer for Medical Image Segmentation},
	author={Athanasios Tragakis and Chaitanya Kaul and Roderick Murray-Smith and Dirk Husmeier},
	booktitle={IEEE/CVF Winter Conference on Applications of Computer Vision},
	pages ={3660--3669},
	year={2023}
	}

@article{Xu2023DUCT,
	title={Cross-teaching with dual uncertainty awareness for semi-supervised medical image segmentation},
	author={Qinglong Xu and Haixing Zhu and Yuan Wang and Zhongjie Shi and Weipeng Liu},
	journal={Multimedia Systems},
	volume={31},
	pages={},
	year={2025}
}

@article{Zhao2025DECSTNet,
  title={Dual Encoder Cross-Shape Transformer Network for Medical Image Segmentation in Internet of Medical Things for Consumer Health},
  author={Zhao, Yawu and Wang, Shudong and Zhang, Yulin and Ren, Yande and Zhang, Yuanyuan and Pang, Shanchen},
  journal={IEEE Transactions on Consumer Electronics},
  year={2025}
}

@article{Zhao2023wranet,
  title={{WRANet}: Wavelet integrated residual attention {U-Net} network for medical image segmentation},
  author={Zhao, Yawu and Wang, Shudong and Zhang, Yulin and Qiao, Sibo and Zhang, Mufei},
  journal={Complex \& Intelligent Systems},
  volume={9},
  number={6},
  pages={6971--6983},
  year={2023}
}

\end{document}